\documentclass[runningheads]{llncs}

\usepackage{eccv}

\usepackage{eccvabbrv}

\usepackage{graphicx}
\usepackage{booktabs}

\usepackage[accsupp]{axessibility}  

\makeatletter
\def\thanks#1{\protected@xdef\@thanks{\@thanks
        \protect\footnotetext{#1}}}
\makeatother

\usepackage{multirow}
\usepackage{bbding}
\usepackage{pifont}

\usepackage[linesnumbered,ruled]{algorithm2e}
\usepackage{threeparttable}

\usepackage{hyperref}

\usepackage{orcidlink}
\usepackage{calligra}
\usepackage{marvosym}

\begin{document}

\title{Can Textual Semantics Mitigate Sounding Object Segmentation Preference?} 

\titlerunning{Sounding Object Segmentation Preference}


\author{Yaoting Wang\inst{1}\inst{\dagger}\orcidlink{0009-0004-5724-5698}\thanks{$^\dagger$ Equal contribution.} \and
Peiwen Sun\inst{2}\inst{\dagger}\orcidlink{0009-0005-3016-8554} \and
Yuanchao Li\inst{3} \orcidlink{0000-0003-4266-2005} \and
Honggang Zhang \inst{2}\orcidlink{0000-0001-8287-6783} \and
Di Hu\textsuperscript{\Letter}\inst{1,4}\orcidlink{0000-0002-7118-6733} 
\thanks{\textsuperscript{\Letter} Corresponding author.}
}
\authorrunning{Y.~Wang et al.}

\institute{
\textsuperscript{1} Gaoling School of Artificial Intelligence, Renmin University of China, China \\
\email{yaoting.wang@outlook.com} \\
\email{dihu@ruc.edu.cn} \\
\textsuperscript{2} Beijing University of Posts and Telecommunications, Beijing, China \\
\email{\{sunpeiwen,zhhg\}@bupt.edu.cn}\\
\textsuperscript{3} University of Edinburgh, Scotland, UK \\
\email{yuanchao.li@ed.ac.uk} \\
\textsuperscript{4} Engineering Research Center of Next-Generation Search and Recommendation\\
}
\maketitle

\begin{abstract}
  The Audio-Visual Segmentation (AVS) task aims to segment sounding objects in the visual space using audio cues.
However, in this work, it is recognized that previous AVS methods show a heavy reliance on detrimental segmentation preferences related to audible objects, rather than precise audio guidance. We argue that the primary reason is that audio lacks robust semantics compared to vision, especially in multi-source sounding scenes, resulting in weak audio guidance over the visual space.
Motivated by the the fact that text modality is well explored and contains rich abstract semantics, we propose leveraging text cues from the visual scene to enhance audio guidance with the semantics inherent in text. 
Our approach begins by obtaining scene descriptions through an off-the-shelf image captioner and prompting a frozen large language model to deduce potential sounding objects as text cues.
Subsequently, we introduce a novel semantics-driven audio modeling module with a dynamic mask to integrate audio features with text cues, leading to representative sounding object features. 
These features not only encompass audio cues but also possess vivid semantics, providing clearer guidance in the visual space.
Experimental results on AVS benchmarks validate that our method exhibits enhanced sensitivity to audio when aided by text cues, achieving highly competitive performance on all three subsets.
Project page: \href{https://github.com/GeWu-Lab/Sounding-Object-Segmentation-Preference}{https://github.com/GeWu-Lab/Sounding-Object-Segmentation-Preference}
\end{abstract}

\section{Introduction}
\label{sec:intro}

In human perception, visual and auditory senses are closely related, and audio can provide vision with additional and dynamic scene information to enhance visual understanding.
In the context of audio-visual understanding tasks, Audio-Visual Localization (AVL) enables the localization of sounding objects within visual scenes using audio references \cite{senocak2018learning,hu2021class}. However, the coarse localization of sounding objects is no longer sufficient to meet the practical demands of complex scene understanding, such as in autonomous driving \cite{van2018susceptibility,zürn2022selfsupervised} and augmented reality \cite{majumder2021move2hear}. To address this challenge, tasks are gradually shifting from bounding-box or rough heat-map localization to finer-grained pixel-level segmented localization, which is known as Audio-Visual Segmentation (AVS).

\begin{figure}[tb]
  \centering
    \includegraphics[width=0.9\linewidth]{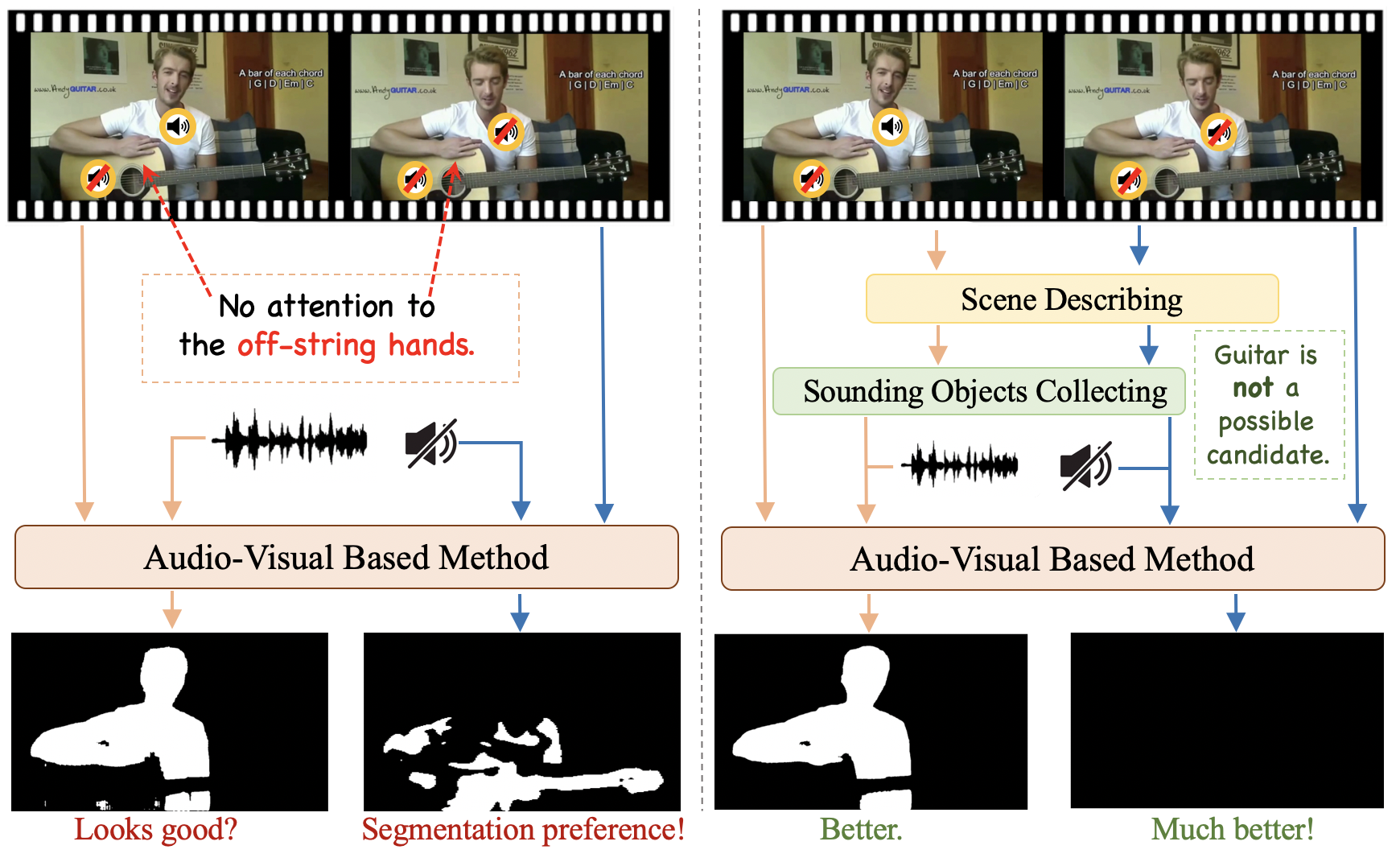}
    \caption{The previous methods \textit{(left)} achieve satisfactory results in receiving normal audio input. However, even when the sound is completely silent, they still segment vast pixels to represent the guitar regarding the segmentation preference of audible objects built during the training. 
    In contrast, our approach \textit{(right)} utilizes a frozen LLM to reason from scene descriptions that the male is not playing the guitar, as his hands are off the strings. Guided by this semantic information, our method produces more precise segmentation with finer audio guidance.}
    \label{fig:real_teaser}
\end{figure}

The existing works on AVS can be broadly categorized into fusion-based methods \cite{zhou2022avs,liu2023audioaware,ling2023hear,huang2023discovering,gao2023avsegformer,li2023catr} and prompt-based methods \cite{wang2023prompting,mo2023av,liu2023annotation}. The former primarily focuses on localizing sounding objects by fusing audio and visual features, while the latter emphasizes constructing effective audio prompts for the visual foundation model. 
However, most of these methods fail to establish compact audio-visual correlations to achieve clear audio guidance in the visual space.
As depicted in the left of \cref{fig:real_teaser}, previous methods could achieve satisfactory results with normal audio input. 
Nevertheless, when the audio is silent, these methods still segment many pixels to represent potential sounding objects based on the \textbf{segmentation preferences} of audible objects\footnote{To be clear, we define audible objects as objects \textbf{\textit{capable}} of producing sound, while sounding objects are defined as objects that are \textbf{\textit{currently}} producing sound.}. 
 
Furthermore, from the experiment with manually muted audio input, we observe that most of the behavior of AVS baselines depends on the segmentation preference established during the training process rather than relying on reliable audio-visual correlation. 
For instance, these models tend to take a shortcut by segmenting out the muted guitars because the audio guidance is weak, and they simply learned with the easy-to-learn visual feature alone (\ie, shortcut learning) that guitars are often associated with sound-emitting during the training phase. 
The unreliable audio-visual correlation can be attributed to two main factors. Firstly, the scarcity of training data for AVS is a significant challenge due to the demanding nature of pixel-level annotation. Secondly, the audio modality itself presents inherent complexity and ambiguity \cite{ling2023hear,huang2023discovering,liu2023audioaware}, especially in scenarios involving multiple audio sources that may be intertwined. 
 
Given the challenges discussed above, our approach aims to build a robust audio-visual correlation by integrating the text modality, which inherently provides meaningful and robust semantic information about objects and their interactions.
\textbf{Firstly}, in the Scene Describing phase, we utilize an off-the-shelf image captioner (LLaVA-1.5 \cite{liu2024visual}) to generate detailed dense captions for scene understanding. 
\textbf{Then}, in the Sounding Objects Reasoning phase, we instruct the frozen large language models (LLMs, \eg, LLaMA2 \cite{touvron2023llama}) as a text cues capturer to collect potential sounding objects from the scene description as text cues. 
Notably, we introduce an elaborate few-shot prompt template to guide the text cues capturer to reason with Chain-of-Thought (CoT) instructions. 
\textbf{After that}, in the Semantics-Driven Audio Modeling (SeDAM) module, we project the audio feature into latent audio features.
However, instead of directly using these latent features to interact with visual features for mask decoding, we leverage these latent features to interact with text cues. 
Specifically, the latent features are attended to and collected to form sounding object features with semantics provided by the text cues through the crossmodal transformer.
We further propose the Prompting Mask Queries with Semantics (PMQS) module to introduce the sounding object features into pre-trained mask queries for later segmentation. \textbf{Finally}, we use simple but effective bottleneck adapters in the Audio-Prompted Decoding phase for better segmentation quality.
The experimental results demonstrate that our proposed method improves audio-visual grounding by incorporating guidance from both audio and text cues. 

Experiments in \cref{sec:audio_control} reveal the advantage of our method, as verified by its increased sensitivity to the changes in audio input compared to networks without enhancing audio guidance with text cues. In summary, our contributions are threefold:

\begin{itemize}
    \item We recognize the behavior of AVS model is influenced by the preference segmentation phenomenon, which can not be solved easily by previous audio-visual based methods. A novel strategy introduces textual semantics as the bridge into the AVS task, establishing a better audio-visual correlation.  

    \item We introduce the SeDAM module to form sounding object features with audio and text cues. We also suggest a PMQS module to prompt the pre-trained mask queries with these sounding object features for sounding object segmentation using a visual foundation model.

    \item Our experiments show that our method can be more sensitive to the audio guidance, manifested in its high sensitivity to the changes in audio input. 
\end{itemize}

\section{Related Works}
\label{sec:formatting}

\subsection{Audio-Visual Localization and Segmentation}

Traditional AVL tasks \cite{senocak2018learning,hu2021class,park2023marginnce} predict the positions of sounding objects using bounding-box or coarse heat-maps through unsupervised learning of audio-visual correlation. 
In recent years, driven by the growing demand for more precise localization in industries such as autonomous driving \cite{van2018susceptibility,zürn2022selfsupervised} and augmented reality \cite{majumder2021move2hear}, the AVL task has evolved towards a finer-grained AVS task that localizes sounding objects at a pixel-level.
The existing AVS works can be broadly categorized into fusion-based \cite{zhou2022avs,liu2023audioaware,ling2023hear,huang2023discovering,gao2023avsegformer,li2023catr,yan2023referred} and prompt-based \cite{wang2023prompting,liu2023annotation,ma2024steppingstones}
The pioneering AVS work \cite{zhou2022avs} is fusion-based and adopts a multi-stage strategy to integrate audio with multi-scale visual features.
Liu \etal \cite{liu2023audioaware} address the issues of inadequate fusion of audio-visual features with an audio-aware query-enhanced transformer (AuTR).  
In comparison, AVSegFormer \cite{gao2023avsegformer} directly decodes the fused feature with the audio query.
Wang \etal \cite{wang2023prompting} innovatively prompt the visual foundation model with audio input, harnessing the abundant visual before achieving generalizable AVS in zero-shot and few-shot scenarios.
However, these methods have not effectively established a reliable audio-visual correlation, preventing the models from perceiving robust audio guidance. On the contrary, our experiments in \cref{sec:audio_control} also reveal that some AVS baselines' competitive performance may stem from segmentation preferences formed during the training.

\subsection{Text Aided Scene Understanding} 

Before the explosion of LLM, textual semantic information had been widely used for visual understanding tasks. Zhan \etal \cite{shi2020improving} and Sharma \etal \cite{sharma2022image} improve Visual Question Answering by using the extra semantic information of image captions, while Hur and Park \cite{hur2023zero} utilize the image captioner to help zero-shot image classification. Wang \etal \cite{wang2024refavs} incorporating text with audio and visual cues to solve reference AVS task.

In recent years, the community has witnessed an emergent interest in strong off-the-shelf capabilities in open-world visual understanding \cite{liu2023llava,liu2024llavanext,liu2023improvedllava,zhang2023videollama}. Notable examples, such as GPT4(V) \cite{yang2023dawn} and LLaVA \cite{liu2023visual} have showcased remarkable linguistic and visual capabilities in zero-shot scene understanding \cite{espejel2023gpt,wu2023gpt4vis} without any training requirement. Innovatively, in the field of AVS, BAVS \cite{liu2023bavs} replaces audio input with textual audio tags from a large pre-trained vision (UniDiffuser \cite{bao2023one}) and audio (BEATs \cite{chen2022beats}) foundation model, establishing a reliable AVS system with foundation knowledge. However, due to the significant gap in class distribution between the pre-training dataset and the AVS dataset, as well as the presence of multi-source scenarios in the AVS task, the extracted audio tags are not ideal to provide accurate semantics. In contrast, we utilize the text modality with robust semantic features to enhance rather than replace the audio guidance and address the segmentation preference issue.

\begin{figure*}[tb]
    \centering
    \includegraphics[width=0.94\linewidth]{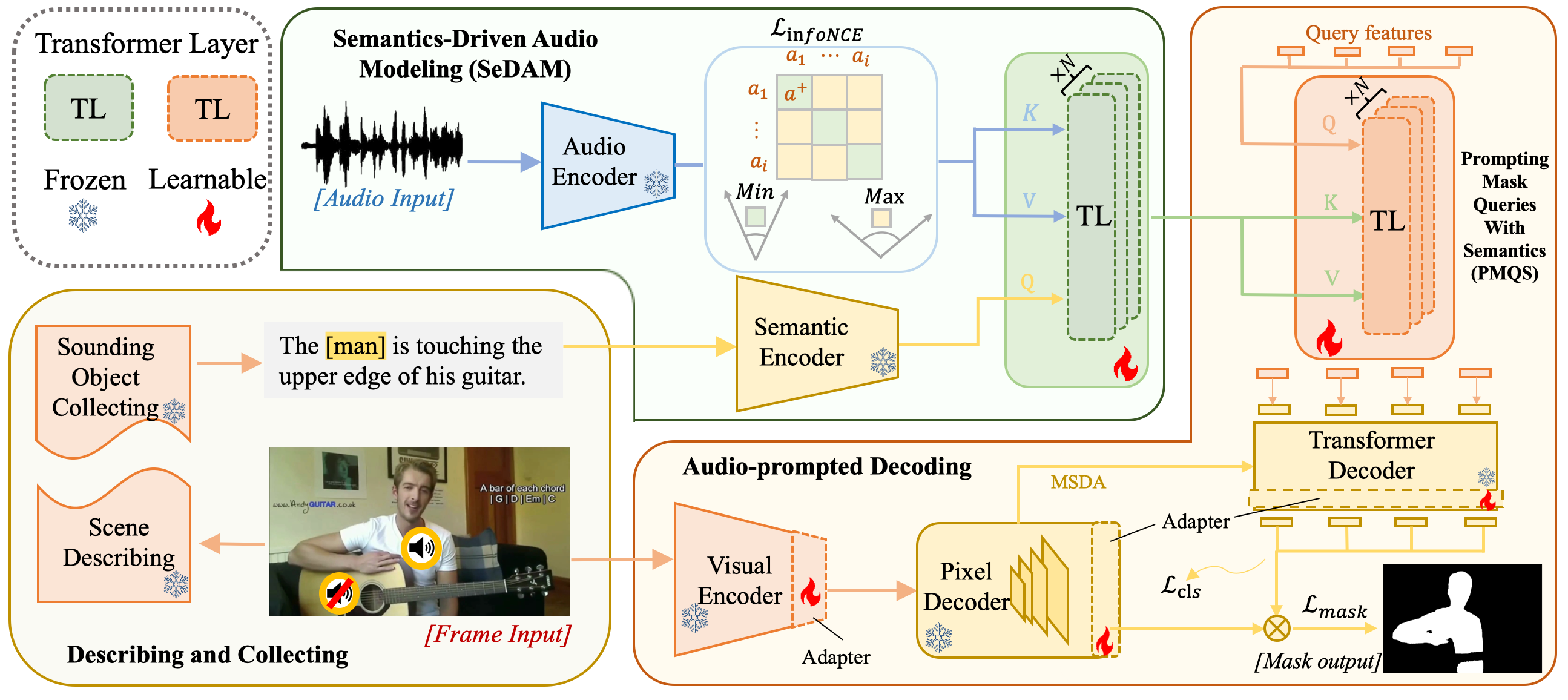}
    \caption{Overall pipeline of the proposed TeSO method. We utilize an off-the-shelf image captioner for dense scene describing and employ a frozen LLM to reason out potential sounding objects as text cues. These semantic text cues are then aggregated with audio features in the SeDAM module to form the sounding object features. Subsequently, we introduce the sounding object features into pre-trained mask queries in the PMQS module. Finally, we use adapters to tune the visual-only mask decoder for AVS in the Audio-prompted Decoding phase. “MSDA” is the multi-scale deformable attention proposed by Zhu \etal. \cite{zhu2020deformable}.
    }
    \label{fig:model}
\end{figure*}

\section{Text-guided Sounding Object Segmentation}
\label{methods}
As shown in \cref{fig:model}, we introduce our \textbf{Te}xt-guided \textbf{S}ounding \textbf{O}bject Segmentation (TeSO) method in this section. Our experiments about muted audio input imply that audio does not exert reasonable guidance over the visual space for some AVS baselines, resulting in the models relying more on the segmentation preferences established during the training process.
In essence, they learn which objects are most likely audible during training and take shortcuts during inference.
To address the above issue, our method aims to enhance the audio-visual correlation by leveraging the text modality, which inherently possesses robust semantic information, to obtain finer-grained audio guidance.
We begin by acquiring detailed scene descriptions through an off-the-shelf image captioner. Subsequently, a frozen LLM works as the text cues capturer, collecting potential sounding objects as text cues from the scene descriptions with CoT instructions.
Finally, we introduce a novel SeDAM module with a dynamic mask to seamlessly integrate audio features and text semantics through a crossmodal transformer, providing finer audio guidance with text semantic cues.

\subsection{Multimodal Representation}
\noindent\textbf{Visual:} Following previous works \cite{zhou2022avs,wang2023prompting}, we sample frames from the video at 1-second intervals. Mask2Former \cite{cheng2021mask2former} serves as our visual foundation model, and we extract visual features $F_V \in \mathbb{R}^{d_V \times H \times W}$ from a pre-trained Swin transformer \cite{dosovitskiy2020image}. We use simple two-layer MLP adapters to tune the visual encoder for better visual representations.
 
\noindent\textbf{Audio:} Similar to the video processing, we split the audio into clips at 1-second intervals. We extract audio features $F_{As}=\{F_{A_1}, F_{A_2}, ..., F_{A_{T}}\} \in \mathbb{R}^{T \times d_{A}}$ using VGGish \cite{gemmeke2017audio,hershey2017cnn}, where $T$ represents the length of the audio in seconds, matching the number of video frames. Therefore, each video frame corresponds to $F_{A} = F_{As}[i]$. We freeze the pre-trained parameters for a better comparison with previous works. 

\noindent\textbf{Text:} For each video frame, we complement the captured text cues into a sentence using a template (``This is a \text{\{\_\}}.'') to extract text features $F_{T} \in \mathbb{R}^{N_T \times d_T}$ with ImageBind \cite{girdhar2023imagebind}, where $N_T$ is the number of text cues. 

\subsection{Text Cues Generation}
 
To obtain more precise text cues, we use a two-stage inference flow. Firstly, we employ an off-the-shelf image captioner to generate dense scene descriptions. Secondly, we ask a text cues capturer to collect potential sounding objects from the generated descriptions.

\subsubsection{Scene Describing.}
\label{sec:scene_under}
\begin{figure}[tb]
    \centering
    \includegraphics[width=0.65\linewidth]{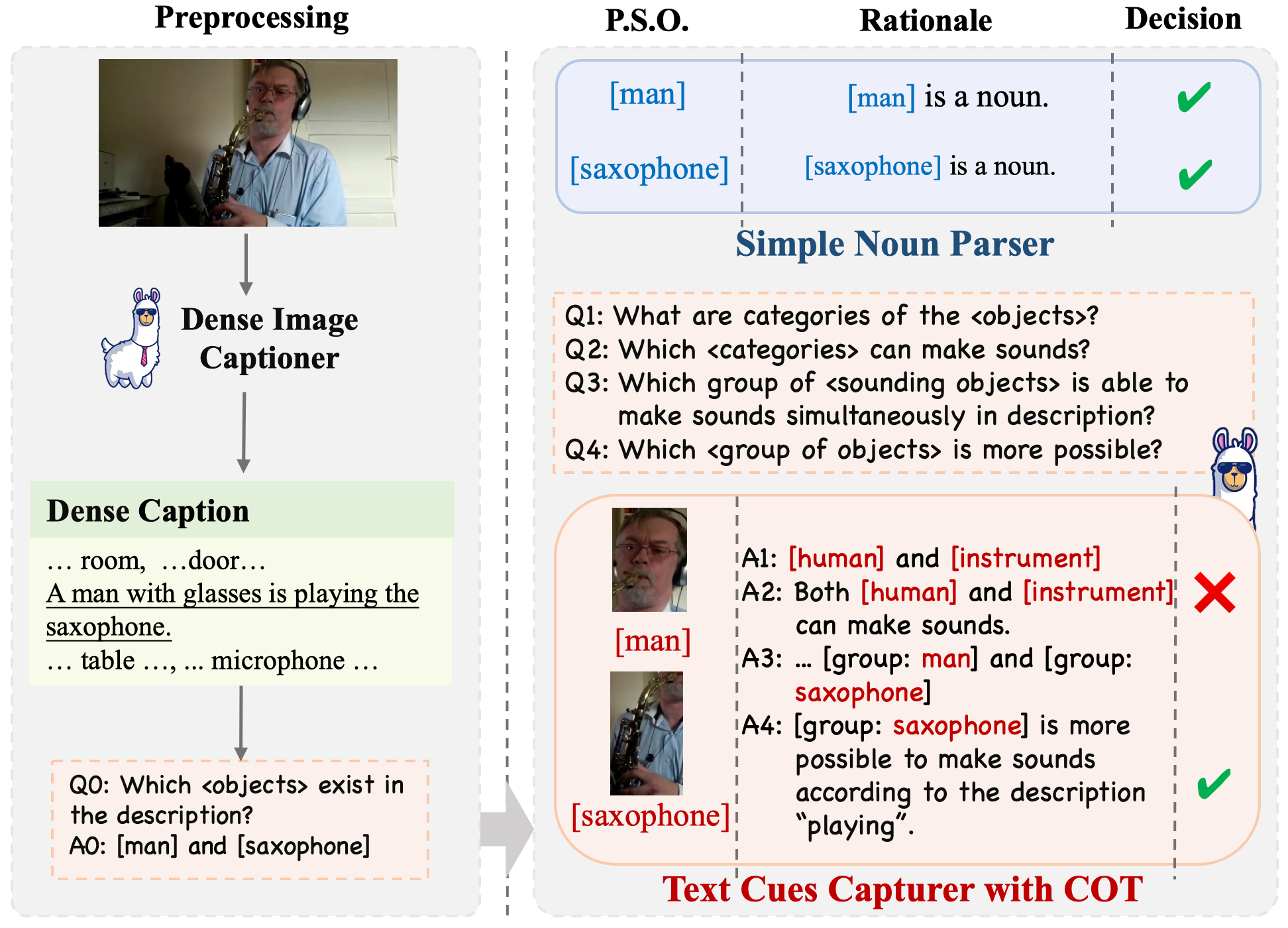}
    \caption{
    ``P.S.O'' stands for ``Potential Sounding Object''. 
    A frozen LLM reasoner works as a text cues capturer by considering the interaction of audible objects. For instance, a person may sing a song while playing the guitar, but he would not sing along with a saxophone. In contrast, a noun parser simply captures any nouns that are present.
    }
    \label{fig:pmp_template}
\end{figure}

As shown in the top-left of \cref{fig:pmp_template}, in this phase, we employ the commonly used and straightforward prompt: \textit{``Please carefully understand the image content and provide as rich a description as possible.''} This instructs the image captioner to generate a detailed description of the visual scene, a task commonly referred to as dense captioning \cite{johnson2016densecap}. 
This process enables the image captioner to produce comprehensive and informative descriptions that not only describe the objects like ``a human and guitar in the image'' but also capture their interactions and relationships.

\subsubsection{Sounding Object Collecting.}
\label{sec:sor}
To extract potential sounding objects from the generated dense descriptions, we design a task-specific CoT prompt template with few-shot demonstrations, as illustrated in \cref{fig:pmp_template}. 
Specifically, we carefully craft examples that cover various scenarios and provide reasoning approaches for each example. 
The process begins by identifying the objects present in the dense scene descriptions, followed by step-by-step common-sense reasoning according to the CoT instructions, as depicted in the lower right of \cref{fig:pmp_template}. 
During the CoT reasoning, the text cues capturer first categorizes the sample objects and assesses whether objects in that category are likely making sounds. Subsequently, the capturer analyzes which group of potential sounding objects can produce sounds simultaneously. Finally, the group with higher confidence is captured as the text cues of length $N_T$ and encoded in $F_T$ for further guidance. 
The right part of \cref{fig:pmp_template} also illustrates why we use the frozen LLM as a text cues capturer instead of using a naive noun parser. It is important to note that the use of LLM is solely for generating high-quality text cues, the LLM itself is not engaged in the training process for a fairness concern.

\subsection{Semantic-Driven Audio Modeling}
After obtaining potential sounding objects, we can enhance the semantics of audio features with these text cues in this SeDAM module.
To allow semantic text cues to capture different aspects of audio features $F_A$, we project $F_A$ into distinct latent features 
$F_{A_L}=\{a_{i}\}_{i=1}^{N_L} \in \mathbb{R}^{N_L \times d_{A_L}}$,
where $N_L$ is the number of latent features.
It is worth noting that, we incorporate an auxiliary loss term $\mathcal{L}_{infoNCE}$ aims at ensuring the distinction among latent audio features. This way, we project audio features into the latent space, where each unique $a_i$ represents a latent audio feature to be captured by semantic text cues, forming clear and loud sounding object features with robust semantics for finer guidance.

To incorporate the semantics of text cues, we employ text cues as the query to attend and combine the relevant latent features through crossmodal multi-head attention, formulated as:

\begin{equation}
\label{eq:cm_a2t_mha}
F_{\widetilde{A}_L} = MultiHead(F_T, F_{A_L}, F_{A_L}),
\end{equation}

\noindent where $F_{\widetilde{A}_L} \in \mathbb{R}^{N_T \times d_A}$ is the attended latent audio features and will be combined with text cues $F_T$ as follows:

\begin{equation}
\label{eq:cm_a2t_collect}
F_{A_T} = FC(LN(F_T + F_{\widetilde{A}_L})),
\end{equation}

\noindent where $F_{A_T} \in \mathbb{R}^{N_T \times d_A}$ is the combined feature with both audio information and semantic text cues, and $FC$ and $LN$ for linear layer and layer normalization. 
Note that the LLM reasoner may generate a different number of text cues, and we either pad with a zero matrix or truncate to $N_T$.
It is noteworthy that we propose task-specific CoT instructions to prompt the text cues capturer to infer more accurate potential sounding objects as the text cues. Additionally, we also introduce a dynamic mask based on attention score to mitigate the potential noisy information. 

We first compute the average pooling attention scores $\zeta \in \mathbb{R}^{N_T}$ for text cues on the attention head and key dimension:
\begin{equation}
\label{eq:cm_a2t_score}
\xi = Pool\left(\frac{{F_T}{F_{A_L}^\text{T}}}{\sqrt{d_{A_L}}}\right),
\end{equation}

\noindent where $\sqrt{d_{A_L}}$ is the scaling factor for dot production. Then we form the attention mask $\mathcal{M}=\{m_1, m_2, m_3, ..., m_{N_T}\}$ as follows:
\begin{equation}
\label{eq:cm_a2t_mask}
\mathcal{M}_{i} = \begin{cases}
0 & \text{if } {\xi}_{i} > 0 \\
-\infty & \text{otherwise}
\end{cases},
\end{equation}
\noindent where $\mathcal{M} \in \mathbb{R}^{N_T}$, will be used as a key padding mask for the PMQS module in the next section to ensure that the attention weight of redundant text cues approaches zero.

\subsection{Prompting Mask Queries with Semantics}
Current segmentation foundation models \cite{cheng2022masked,cheng2021mask2former,kirillov2023segany} adopt transformer-based decoder with mask queries as the mainstream approach. 
To prompt such foundation models, we propose the PMQS module, aiming to introduce semantic information to mask query features $F_Q=\{q_1, q_2, ..., q_{N_Q}\}$, where $F_Q \in \mathbb{R}^{N_Q \times d_V}$, and $N_Q$ is the number of mask queries in crossmodal attention. 
We formulate this updated expression as follows:

\begin{equation}
\label{eq:cm_mq_attn}
{\phi}_{ca}{(F_{Q}, F_{A_T}, F_{A_T})} = Softmax\left(\frac{F_{Q}{F_{A_T}^\text{T}}}{\sqrt{d_{A_T}}} + \mathcal{M}\right)F_{A_T},
\end{equation}
\noindent where $\mathcal{M}$ is added to the attention score to mask the redundant text cues for each query. Then we obtain the updated mask query features $\widetilde{F}_Q \in \mathbb{R}^{N_Q \times d_V}$ as:

\begin{equation}
\label{eq:cm_mp_add}
\widetilde{F}_Q = {F}_Q + {\phi}_{ca}{(F_{Q}, F_{A_T}, F_{A_T})}.
\end{equation}

\subsection{Audio-Prompted Decoding}



As shown on the right side of \cref{fig:model}, the updated mask query features $\widetilde{F}_Q$ are forwarded into the transformer-based decoder, generating mask queries $M_Q \in \mathbb{R}^{N_Q \times H \times W}$ and class queries $C_Q \in \mathbb{R}^{N_Q \times N_C}$, where $N_C$ is the number of class queries pre-defined in the segmentation foundation model. Subsequently, we can obtain the final segmentation $M_{pred} \in \mathbb{R}^{N_C \times H \times W}$:
\begin{equation}
\label{eq:final_seg}
M_{pred} = Einsum(C_Q, \widetilde{F}_Q),
\end{equation}
\noindent where $Einsum(\cdot)$ is the Einstein Summation Convention.

Notably, to enhance the quality of the segmentation, we use two-layer bottleneck MLP adapters \cite{houlsby2019parameter} to lightly fine-tune the output projection of the following attentions: a) Swin Window Attention (W-MSA) 
\cite{liu2021swin}, b) Deformable Attention in pixel decoder \cite{zhu2020deformable} and c) Masked Attention in transformer decoder \cite{cheng2021mask2former}.

\subsection{Learning Objectives}
\noindent\textbf{Segmentation Loss.} 
During the model training process, we employ the binary cross-entropy loss and dice loss to optimize the mask quality: 
\begin{equation}
\label{eq:loss_seg_mask}
    \mathcal{L}_{mask} = \lambda_{bce} \cdot \mathcal{L}_{bce} + \lambda_{dice} \cdot \mathcal{L}_{dice}.
\end{equation}
Further, we adopt the mask classification loss \cite{cheng2022masked} to compose the final segmentation loss:
\begin{equation}
\label{eq:loss_seg}
    \mathcal{L}_{seg} = \mathcal{L}_{mask} + \lambda_{cls} \cdot \mathcal{L}_{cls}.
\end{equation}

\noindent\textbf{Latent Component Loss.} This loss term aims to explicitly guide the model to learn distinct representations for each latent audio component, helping prevent redundancy and overlapping information across the latent components. 
\begin{equation}
\mathcal{L}_{info\textit{NCE}} = -\log\left(\frac{{\exp({{sim}}(a, a^+))}}{{\sum_{i=1}^{N_L}\exp({{sim}}(a, a_i))}}\right),
\end{equation}
\noindent where $a^+$ and $sim(\cdot)$ denote the positive example and the cosine similarity.

\noindent\textbf{Total Loss}
The final loss function is composed of the weighted sum of the two kinds of losses above:
\begin{equation}
\label{eq:tot_loss}
\mathcal{L} = \mathcal{L}_{seg} +{\lambda}_{info\textit{NCE}} \cdot \mathcal{L}_{info\textit{NCE}}.
\end{equation}




\section{Experiments}
\label{sec:experiment}


\subsection{Implementation Details}
\label{sec:details}

\textbf{Dataset.} Our proposed method is evaluated on the AVS Benchmarks \cite{zhou2022avs}, which contains three subsets. Firstly, the single-source subset (V1S) contains 4932 videos over 23 categories, covering various sounds, such as humans, animals, vehicles, and musical instruments. For videos in the training split of this subset, only the first sampled frame is annotated. Secondly, the multi-source subset (V1M) contains 424 videos that include two or more categories from V1S, and all sounding objects are visible in the frames. Finally, the AVSBench-semantic (AVSS) subset, an extension of V1S and V1M, contains 12,356 videos. These newly collected videos are trimmed to 10 seconds, differing from the 5 seconds in V1S and V1M.


\noindent\textbf{Setting.} We conduct training and evaluation using the VGGish backbone pretrained on Youtube-8M \cite{abu2016youtube} and Swin-base Transformer backbone pretrained on semantic-ADE20K \cite{zhou2019semantic} by Mask2Former \cite{cheng2021mask2former}. 
The number $N$ of the crossmodal transformer layers is set to $4$, and the parameters $\lambda_{bce}, \lambda_{dice},\lambda_{cls},\lambda_{info\textit{NCE}}$ in the loss are set to $5,5,2,1$. The AdamW optimizer is adopted with a learning rate of 1e-4 for the visual encoder adapters and 1e-3 for other learnable parameters, and the training epoch is roughly set to 60.

\noindent\textbf{Metrics.} To conduct a comprehensive evaluation of our model, we carry out tests using mean Intersection over Union (mIoU) and F-score as the performance metrics, following previous works \cite{zhou2022avs,gao2023avsegformer}. 

\subsection{Comparison Results}
\label{sec:avs_benchmarks}

When compared to methods that do not incorporate text as additional semantic information, our method, which leverages captioning and reasoning, demonstrates strong competitiveness. As illustrated in \cref{tab:full-avs}, our method achieves comparable results across all V1S, V1M, and AVSS-binary subsets, showcasing average performance improvements of up to 1.25\% in mIoU and 2.7\% in F-score, respectively. Furthermore, qualitative analysis indicates that our predicted masks exhibit superior quality, as demonstrated in sub-figure (1) of Fig. \ref{fig:audio_control} (more examples in supplementary materials). 

\begin{table*}[tb]
    \centering
    \scalebox{0.78}{
    \begin{threeparttable}
    \caption{Performance on AVS-Benchmarks. The \underline{underscore} is used to indicate suboptimal results.
    BAVS uses large pre-trained visual and audio foundation models to replace audio input with text tag. 
    In comparison, TeSO (ours) enhances rather than replaces audio guidance based on the robust semantics of text. 
    It is evident that our method has a good F-score result while ensuring mIoU.}
    \begin{tabular}{lcccccccc}
        \hline
        \multirow{2}{*}{Method} & \multirow{2}{*}{Audio-backbone} & \multirow{2}{*}{Visual-backbone} & \multicolumn{2}{c}{V1S} & \multicolumn{2}{c}{V1M} & \multicolumn{2}{c}{AVSS-binary}  \\ 
         &  &  & mIoU(\%) & F-score & mIoU(\%) & F-score & mIoU(\%) & F-score \\ 
    
        \hline
        AVSBench \cite{zhou2022avs}& VGGish & PVT-v2 & 78.70 & 0.879 & 54.00 & 0.645 &  62.45 & 0.756 \\ 
        AVSegFormer \cite{gao2023avsegformer}& VGGish & PVT-v2 & 82.06 & 0.899 & 58.36 & 0.693 &  64.34 & 0.759 \\ 
   
        AVSC \cite{liu2023audiovisual}& VGGish & PVT-v2 & 81.29 & 0.886 & 59.50 & 0.657 & - & - \\

        AVS-BG \cite{hao2023improving} & VGGish & PVT-v2  & 81.71 & \underline{0.904} & 55.10 & 0.668   & - & - \\ 

        AQFormer \cite{huang2023discovering} & VGGish & PVT-v2  & 81.60 & 0.894 & 61.10 & 0.721   & - & - \\ 

        $^\dagger$CATR \cite{li2023catr} & VGGish & PVT-v2 & 81.40 & 0.896 & 59.00 & 0.700 & - & - \\

        AV-SAM \cite{mo2023av} &  ResNet18 & ViT-Base & 40.47 & 0.566 & - & - & - & -\\

        SAMA \cite{liu2023annotation} & VGGish & ViT-Huge  & 81.53 & 0.886 & 63.14 & 0.691   & - & - \\ 
        
        GAVS \cite{wang2023prompting} & VGGish & ViT-Base & 80.06 & 0.902 & 63.70 & \underline{0.774}   & \underline{67.70} & \underline{0.788} \\

        AuTR \cite{liu2023audioaware} & VGGish & Swin-Base & 80.40 & 0.891 & 56.20 & 0.672 & - & - \\

        MUTR \cite{yan2023referred} & VGGish & Video-Swin-Base & 81.60 & 0.897 & \underline{64.00} & 0.735 & - & - \\

        \hline 


        BAVS \cite{liu2023bavs} & $^\ddagger$BEATs & Swin-Base & \underline{82.68} & 0.898 & 59.63 & 0.659   & 55.45 & 0.640 \\


        TeSO (ours) & VGGish & Swin-base & \textbf{83.27} & \textbf{0.933} & \textbf{66.02} & \textbf{0.801}   & \textbf{68.53} & \textbf{0.813} \\ 
        
        \hline
    \end{tabular}
    \begin{tablenotes}
        \small 
        \item[$\dagger$]: To make a fair comparison, the results of CATR here are without supplemented annotation of the training set.
        \item[$\ddagger$]: BEATs is a stronger audio backbone, trained on AudioSet dataset with 90M parameters. In comparison, VGGish is trained on Youtube-8M with 70M parameters.
    \end{tablenotes}
    \end{threeparttable}
    }
    \label{tab:full-avs}
\end{table*}

It is necessary to mention that our proposed TeSO does not show significant improvements on V1S over certain existing methods, such as BAVS \cite{liu2023bavs}, which also utilizes text semantic information. Apart from marginal effects, this phenomenon can also be attributed to the singular sound source and simplistic nature of the data scenarios present in V1S. On V1M, however, our performance significantly surpasses that of other methods, including BAVS.


In addition to the AVS task, we also conduct experiments on the Audio-Visual Semantic Segmentation (AVSS) task, as shown in \cref{tab:avss}. Our model exceeds the best performance in previous models by 5.37\% on mIoU, which further demonstrates that our model is capable of establishing a better audio-visual correlation.

\begin{table*}[tb]
  \begin{minipage}[t]{0.47\linewidth} 
    \centering
\caption{Performance on AVSS dataset. Our method showcases a substantial improvement in the semantic subset, achieved by leveraging robust text semantics to enhance audio guidance.}
\scalebox{0.85}{
\begin{tabular}{lccc}
\hline
\multirow{2}{*}{Method} & \multirow{2}{*}{Backbone} & \multicolumn{2}{c}{AVSS} \\
 & & mIoU(\%) & F-Score \\ \hline
AVSBench & PVT-v2 & 29.80 & 0.352 \\
CATR & PVT-v2 & 32.80 & 0.385 \\ \hline
BAVS & Swin-base & 33.59 & 0.375 \\ 
TeSO (Ours) & Swin-base &  \textbf{38.96} & \textbf{0.451} \\ \hline
\end{tabular}
}
\label{tab:avss}
  \end{minipage}
  \hfill
  \begin{minipage}[t]{0.47\linewidth}
    \centering
\centering
\caption{Potential sounding objects recognition (REC) performance at different model stages and the corresponding segmentation results. Experiments are conducted on AVS-V1M test set.}
\scalebox{0.85}{
\begin{tabular}{l|cc|c}
\hline
\multirow{2}{*}{Method}  & REC$_{(\#1)}$ & REC$_{(\#2)}$ & SEG \\
   & AP(\%) & AP(\%) & mIoU(\%) \\ \hline
BAVS & 24.22 & 79.69 & 59.63 \\
TeSO$_{p}$ (ours) & 77.96 & 78.33 & 63.87  \\  
TeSO$_{n}$ (ours) & 77.82 & 78.27 & 63.31  \\  
TeSO (ours) & \textbf{80.41} & \textbf{80.95} & \textbf{66.02} \\
\hline
\end{tabular}
}
\label{tab:pso_cls}
  \end{minipage}
\end{table*}

\subsection{Bottlenecks and Fairness}

Our method (TeSO) collects text cues from dense image descriptions at REC$_{\#1}$ (80.41\%) and a filter with a dynamic mask at REC$_{\#2}$ (80.95\%), indicating a robust startup for the text-guided AVS framework, in comparison with another two-stage model, BAVS. 

In addition, we develop several variants of TeSO to investigate its performance sources and bottlenecks, as well as its fairness in comparison with other methods.
TeSO$_{p}$ denotes the variant using one-shot prompts, while TeSO$_{n}$ denotes the variant employing a noun parser instead of a frozen LLM for text cues collection. 
As shown in \cref{tab:pso_cls}, the results demonstrate that both the number (shots) of demonstrations and the method of text cues collection significantly impact the model's ability to accurately identify audio semantics and perform segmentation (SEG). 
Thus, utilizing few-shot prompts and off-the-shelf foundation models are both essential for ensuring the acquisition of high-quality text cues (\ie, the bottleneck).

\textbf{However}, we would like to emphasize that, even without using LLM to capture text cues, our method variant (TeSO$_n$) still fairly achieves competitive performance (63.31\% mIoU) compared with BAVS (59.63\% mIoU) and state-of-the-art method GAVS (63.70\% mIoU) in the complex multi-source scenario (AVS-V1M), and significantly outperforms other methods. 
This suggests that leveraging text-enhanced audio for visual control with robust semantics is a viable and promising direction.

\subsection{Audio Control}
\label{sec:audio_control}
\begin{figure*}[tb]
  \centering
   \includegraphics[width=0.82\linewidth]{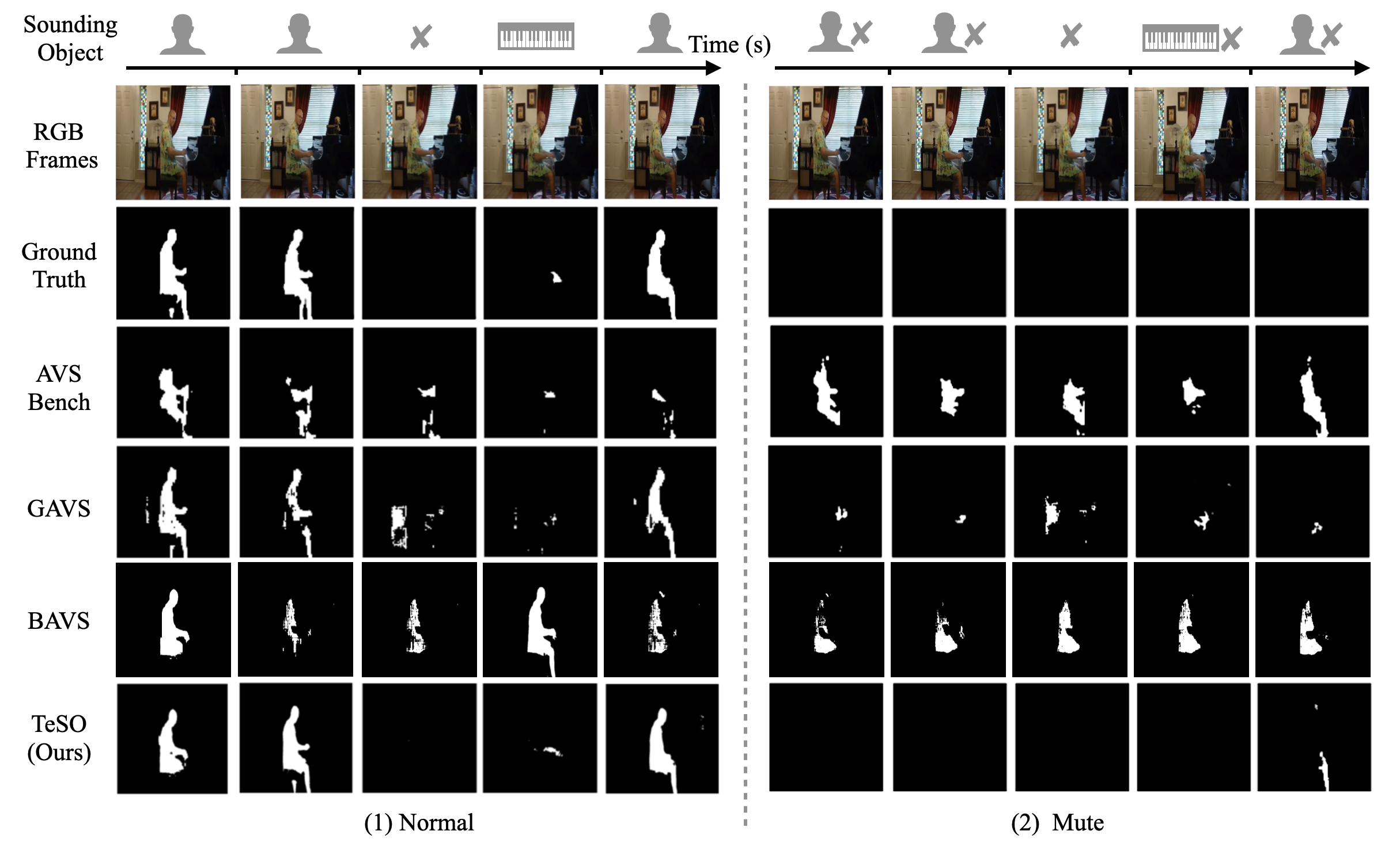}
   \caption{Examples of the impact of normal audio input and all-mute audio on popular methods. In normal scenarios, our method shows better masks than previous methods. In all-mute scenarios, our approach exhibits strong sensitivity towards audio inputs, as it is capable of generating blank masks for silent audio clips.}
   \label{fig:audio_control}
\end{figure*}

Based on our observations, we notice a distinct shape of the predicted mask in sub-figure (2) of Fig. \ref{fig:audio_control} when some popular AVS baselines are used for inference. Interestingly, this shape remains distinguishable even when the inference is conducted with muted audio. This suggests that the network can predict rough masks without relying on audio input, indicating that the audio-visual grounding is not effectively guided by audio cues. 
As a result, this undesirable phenomenon can be attributed to the segmentation preference, which is a result of these models lacking proper audio control.

To evaluate the influence of audio control on segmentation, various models are subjected to inference using muted audio and noise-only audio (at 10 dB and 40 dB) instead of the corresponding audio signals. It should be noted that the visual features remained unchanged throughout the evaluation. 
As shown in Tab. \ref{tab:audio_control}, the disparity $\Delta_{m}$ (and $\Delta_{f}$) refers to the difference in mIoU (and F-score) values before and after the audio substitution is applied as an evaluation metric for the sensitivity of audio changes.
A larger drop in mIoU and F-score indicates a greater degree of audio control over the visual space. 
In addition to demonstrating strong segmentation capabilities in the AVS task, TeSO (ours) exhibits the highest level of attenuation, reaching up to 47.4\%, when subjected to muted and noise-only audio. Additionally, qualitative analysis revealed that the predicted masks remained predominantly blank, as depicted in sub-figure (2) of \cref{fig:audio_control} (more examples in supplementary materials). This observation indicates that our model has better audio guidance over the visual space, enabling it to play a crucial role in audio-visual grounding.

\begin{table*}[tb]
\centering
\caption{Impact on mIoU and F-score by muted audio or noise-only audio on AVS-V1M test set. TeSO (ours) exhibits the most intensive attenuation up to 48.8\%. It is noteworthy that the other models still generate ``satisfactory'' results even in silent or purely noisy environments. This observation suggests that the models might rely more on the segmentation preferences formed during the training rather than truly learning efficient audio guidance.}

\scalebox{0.72}{
\begin{tabular}{lcccccccccccc}
\hline

& \multicolumn{2}{c}{Mute} & \multicolumn{2}{c}{WGN-10dB} & \multicolumn{2}{c}{WGN-40dB} \\

{Method} & mIoU\% (\textcolor{blue}{$\Delta_{m}$\%}) &  F-score (\textcolor{blue}{$\Delta_{f}$\%}) &  mIoU\% (\textcolor{blue}{$\Delta_{m}$\%}) &  F-score (\textcolor{blue}{$\Delta_{f}$\%}) & mIoU\% (\textcolor{blue}{$\Delta_{m}$\%}) & F-score (\textcolor{blue}{$\Delta_{f}$\%}) \\ \hline

{AVS-Bench} &  49.15 (\textcolor{blue}{$\mathtt{6.6}$}) &  0.622 (\textcolor{blue}{$\mathtt{4.9}$}) &  46.76 (\textcolor{blue}{$\mathtt{11.1}$})&  0.618 (\textcolor{blue}{$\mathtt{5.5}$}) & 47.46 (\textcolor{blue}{$\mathtt{9.8}$}) &  0.610 (\textcolor{blue}{$\mathtt{6.7}$})\\

{GAVS} &  42.53 (\textcolor{blue}{$\mathtt{29.2}$}) &  0.695 (\textcolor{blue}{$\mathtt{8.4}$}) & 41.51 (\textcolor{blue}{$\mathtt{34.8}$}) &  0.692 (\textcolor{blue}{$\mathtt{10.6}$}) &  41.87 (\textcolor{blue}{$\mathtt{30.3}$}) &  0.696 (\textcolor{blue}{$\mathtt{8.4}$}) \\

{BAVS} &  45.06 (\textcolor{blue}{$\mathtt{26.1}$}) &  0.644 (\textcolor{blue}{$\mathtt{5.8}$}) & 43.27 (\textcolor{blue}{$\mathtt{27.4}$}) &  0.601 (\textcolor{blue}{$\mathtt{8.8}$}) &  42.93 (\textcolor{blue}{$\mathtt{31.6}$}) &  0.595 (\textcolor{blue}{$\mathtt{9.7}$}) \\

{TeSO (ours)} &  37.89 (\textcolor{blue}{$\mathtt{\textbf{42.6}}$}) &  0.671 (\textcolor{blue}{$\mathtt{\textbf{16.2}}$}) & 34.20 (\textcolor{blue}{$\mathtt{\textbf{49.6}}$}) & 0.661 (\textcolor{blue}{$\mathtt{\textbf{17.5}}$}) & 34.74 (\textcolor{blue}{$\mathtt{\textbf{47.4}}$}) &  0.656 (\textcolor{blue}{$\mathtt{\textbf{18.1}}$})  \\

\hline
\end{tabular}
}
\label{tab:audio_control}
\end{table*}

\subsection{Ablation Study}

As shown in Tab. \ref{tab:ablation_control}, we conduct ablation experiments targeting the designed TeSO modules and the guidance from different modalities. 
To explore the guiding ability, we progressively remove either the audio or text guidance for TeSO. 
The experiments indicate that control from both audio and text inputs for visual space can establish a strong audio-visual correlation. 
Furthermore, by utilizing our proposed TeSO, we achieve enhanced audio guidance and reach a 66.02\% mIoU while incorporating precise semantic information from the text modality. 
Moreover, through the utilization of our proposed TeSO method that incorporates robust semantic information from the text modality, we achieved improved audio guidance and attained a mIoU of 66.02\%.

\begin{table*}[tbp]
  \begin{minipage}[t]{0.44\linewidth} 
    \centering
     \caption{Ablation study on AVS-V1M. Our method benefits from using both audio and text as the guiding information for sounding objects.}

    \centering
    \scalebox{0.8}{
    \begin{threeparttable}
    \begin{tabular}{lcc}
\hline
Control   & mIoU(\%) & F-score \\ \hline
TeSO  & \textbf{66.02}    & \textbf{0.801}  \\ 
(-) $\mathcal{L}_{infoNCE}$ & 65.34 & 0.793 \\
(-) Dynamic Mask & 64.78 & 0.781 \\
(-) PMQS & 62.90 & 0.774 \\
(-) SeDAM & 62.22 & 0.760 \\ \hline
(-) Text  & 62.28    & 0.768   \\ 
(-) Audio & 42.11    & 0.692   \\
(-) Text \& Audio  & 37.35    & 0.656   \\
\hline
 \end{tabular}
 \end{threeparttable}
 }
    
 \label{tab:ablation_control}
  \end{minipage}
  \hfill
  \begin{minipage}[t]{0.5\linewidth}
    \centering
\centering
\caption{Comparison of our proposed baseline and TeSO. The baseline method employs LLM to filter the predicted results but does not enhance the audio guidance with text cues. Experiments are conducted on AVS-V1M test set. }
\scalebox{0.8}{
\begin{tabular}{lccc}
\hline
   & Normal & Mute & WGN-40dB \\ 
   & mIoU\% & mIoU\% $(\textcolor{blue}{{\Delta}_m\%})$ & mIoU\% $(\textcolor{blue}{{\Delta}_m\%)}$ \\ \hline
Baseline & 64.15 & 43.37 (\textcolor{blue}{32.4}) & 43.09 (\textcolor{blue}{32.8}) \\
TeSO & \textbf{66.02} & 37.89 (\textcolor{blue}{\textbf{42.6}}) & 34.74 (\textcolor{blue}{\textbf{47.4}}) \\
\hline
\end{tabular}
}
\label{tab:baseline}

  \end{minipage}
\end{table*}

As shown in \cref{tab:baseline}, we also implement a baseline that directly filters the inferred semantic labels during the model evaluation process by intersecting them with the potential sounding object collected by the frozen LLM capturer. This implies that no text cues are utilized to enhance the audio guidance for better control over the visual space. As expected, the sensitivity (${\Delta}_m$) of model performance to changes in audio input has decreased compared with TeSO, indicating the effectiveness of our method.

\section{Discussion}
 
\noindent\textbf{Noisy text cues.} Besides dense image captions for collecting semantic text cues, we also constructed a dynamic mask in \cref{eq:cm_a2t_mask} to further select effective semantics based on audio correspondence. We use audio recognition results to determine the accuracy of text cues. Please refer to \cref{tab:pso_cls} to compare the audio recognition results based on text cues collection stage (REC$_{\#1}$) and dynamic mask filtering stage (REC$_{\#2}$). Our approach achieves higher audio recognition performance, compared with using the audio foundation model in BAVS. This indicates the reliability and accuracy of our strategy for obtaining text cues.

\noindent\textbf{Prompt with CoT.}
As shown in the \cref{fig:pmp_template}, we utilize task-specific prompt templates with multiple scenario demonstrations and CoT instructions to assist the frozen LLM in collecting more accurate potential sounding objects as text cues. For the detailed prompt templates with CoT instructions and related experimental information, please refer to the Prompt Template part and Text Assistance part in the supplementary materials.

\noindent\textbf{Versatility.} We explore an alternative visual foundation model like the Segment Anything Model \cite{kirillov2023segany} for segmentation.
Conducting experiments using a ViT-based backbone with minimal adjustments, our model achieves an impressive 65.36\% mIoU on V1M, still getting comparable performance. This demonstrates the versatility and effectiveness of our model. Additional experiment details and analysis can be found in the supplementary materials.
 
\section{Conclusion}
This study represents a pioneering effort to incorporate text cues into the AVS task. 
Unlike previous approaches that solely rely on audio assistance, our TeSO method establishes a stronger audio-visual correlation by harnessing the robust semantics unique to the text modality
By integrating CoT instructions and common-sense reasoning, we enhance the ability to capture potential sounding objects from detailed scene descriptions, thereby optimizing the utilization of text cues. 
Furthermore, we have developed a dynamic mask based on attention scores to effectively handle noisy information present in the text cues.
Our comprehensive experiments highlight the effectiveness of our approach, demonstrating comparative performance across all AVS Benchmarks. 
In particular, more observation experiments validate that enhancing audio with text cues grants superior audio guidance. 
In summary, our work provides a new direction for building reliable audio-visual correlation with textual semantics, and we hope that it can help the community construct more robust audio-visual systems.



\section*{Acknowledgements}
This research was supported by National Natural Science Foundation of China (NO.62106272), and Public Computing Cloud, Renmin University of China.

%
%
\bibliographystyle{splncs04}
\bibliography{main}

\appendix
\renewcommand\thefigure{\Alph{section}\arabic{figure}}
\renewcommand\thetable{\Alph{section}\arabic{table}} 

\title{Can Textual Semantics Mitigate Sounding Object Segmentation Preference? \\ (Supplementary Material)} 

\titlerunning{Sounding Object Segmentation Preference}

\author{Yaoting Wang\inst{1}\inst{\dagger}\orcidlink{0009-0004-5724-5698}\thanks{$^\dagger$ Equal contribution.} \and
Peiwen Sun\inst{2}\inst{\dagger}\orcidlink{0009-0005-3016-8554} \and
Yuanchao Li\inst{3} \orcidlink{0000-0003-4266-2005} \and
Honggang Zhang \inst{2}\orcidlink{0000-0001-8287-6783} \and
Di Hu\textsuperscript{\Letter}\inst{1,4}\orcidlink{0000-0002-7118-6733} 
\thanks{\textsuperscript{\Letter} Corresponding author.}
}
\authorrunning{Y.~Wang et al.}

\institute{
\textsuperscript{1} Gaoling School of Artificial Intelligence, Renmin University of China, China \\
\email{yaoting.wang@outlook.com} \\
\email{dihu@ruc.edu.cn} \\
\textsuperscript{2} Beijing University of Posts and Telecommunications, Beijing, China \\
\email{\{sunpeiwen,zhhg\}@bupt.edu.cn}\\
\textsuperscript{3} University of Edinburgh, Scotland, UK \\
\email{yuanchao.li@ed.ac.uk} \\
\textsuperscript{4} Engineering Research Center of Next-Generation Search and Recommendation\\
}

\maketitle

\section{Detailed Prompts}
\begin{figure*}[!htbp]
  \centering
   \includegraphics[width=0.78\linewidth]{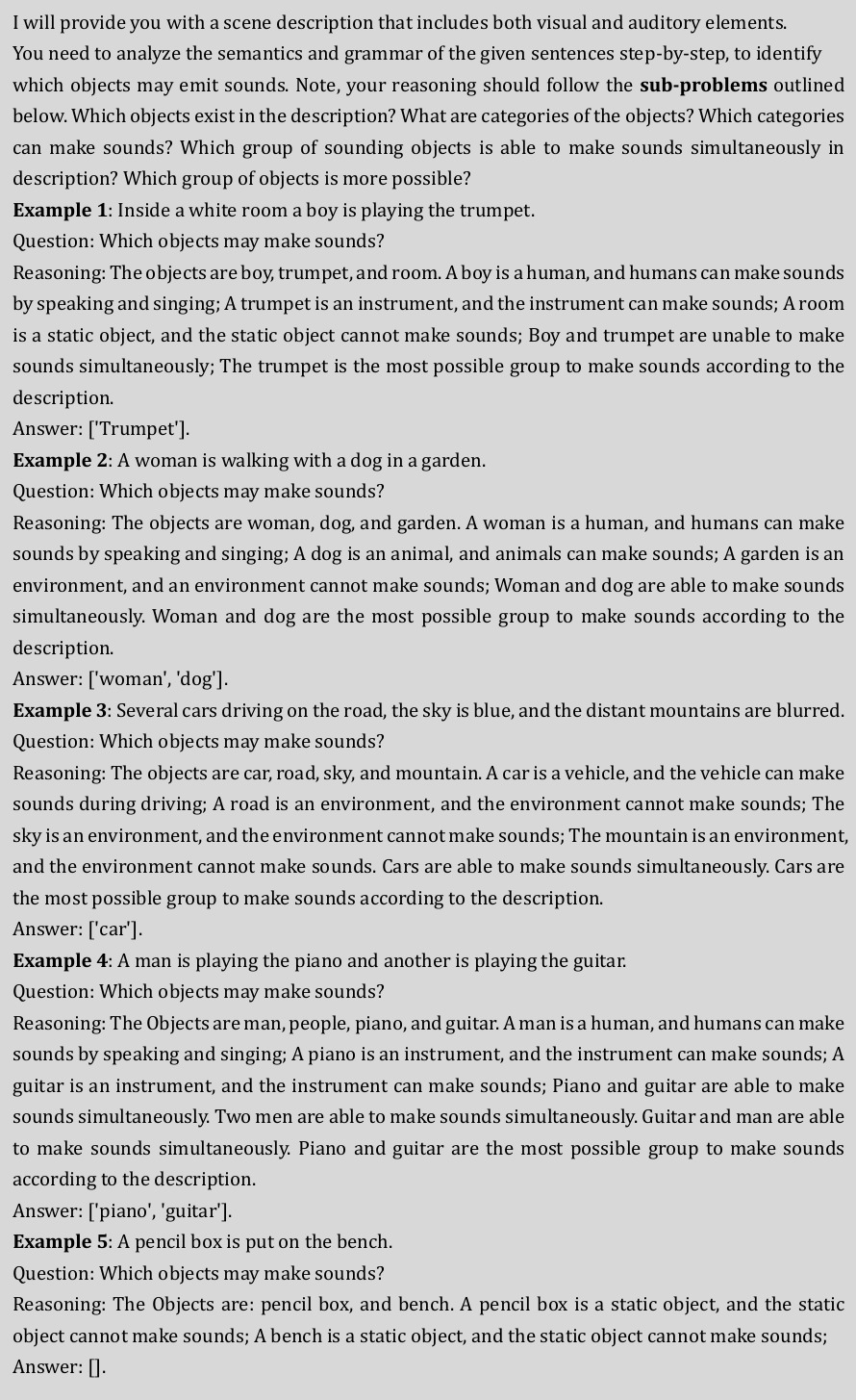}
   \caption{Few-shot prompt with CoT instructions. We feed these prompts to LLMs before each time we generate the reasoning results. This template generates the best result of \cref{tab:prompt_template}.}
   \label{fig:prompts}
\end{figure*}

Our detailed prompt template with CoT instructions for LLMs to reason the potential sounding objects is shown in \cref{fig:prompts}. To enhance the comprehensiveness of instruction examples, we generate examples based on various quantitative and scenario settings. These include scenarios ranging from no possible sounding object to multiple sounding objects, as well as simple sounding scenarios with static objects to complex sounding scenarios with multiple sound sources.


\section{Qualitative Segmentation Results}
\begin{figure*}[tb]
  \centering
   \includegraphics[width=0.72\linewidth]{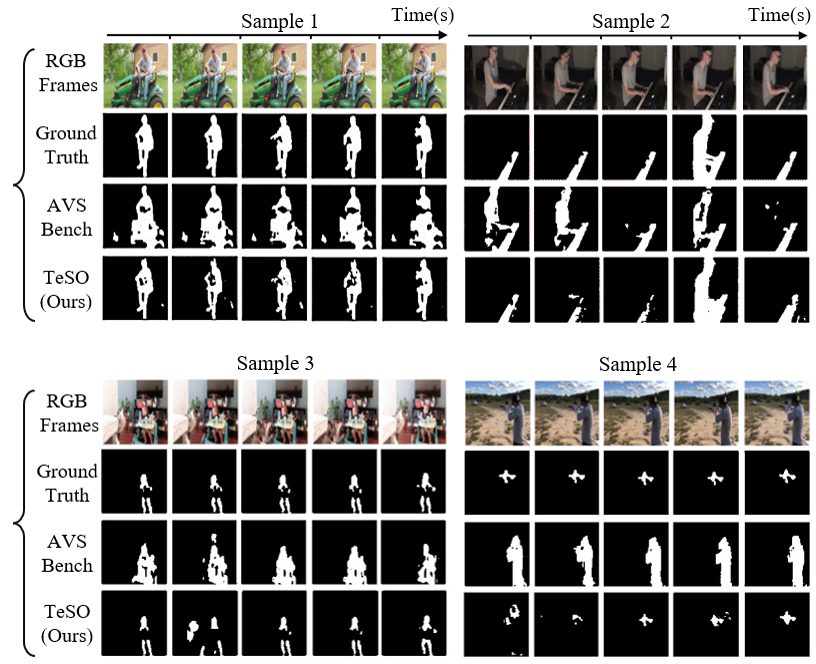}
   \caption{Qualitative comparison on AVS task. TeSO shows brilliant audio-visual grounding compared with the baseline.}
   \label{fig:case_appendix}
\end{figure*}

\begin{figure*}[tb]
  \centering
   \includegraphics[width=0.88\linewidth]{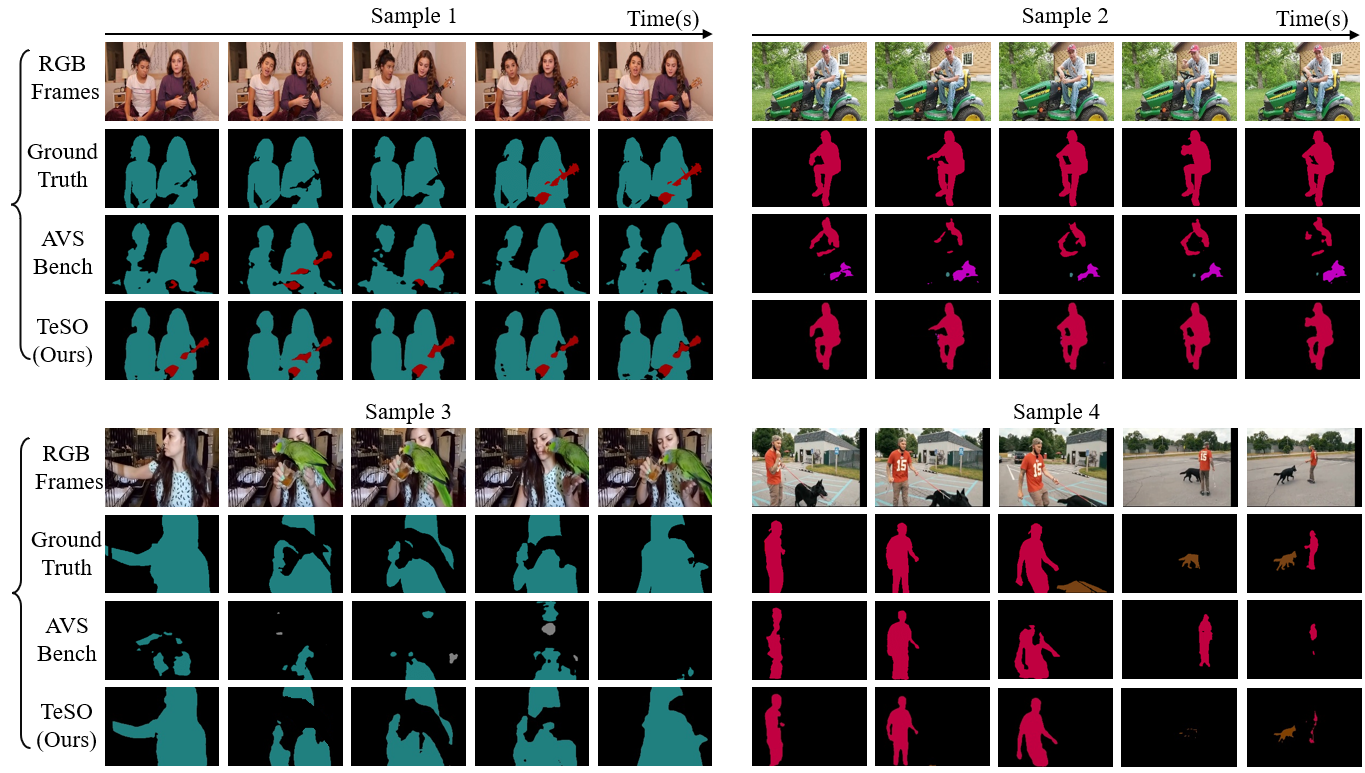}
   \caption{Qualitative comparison on AVSS task. TeSO has better semantic perception capability compared with other method.}
   \label{fig:case_avss_appendix}
\end{figure*}

To demonstrate the effectiveness of our model, we showcase its performance under various conditions by presenting qualitative segmentation examples. These include the results on the AVS task, depicted in \cref{fig:case_appendix}, as well as the results on the AVSS task, shown in \cref{fig:case_avss_appendix}. Due to the unavailability of published work on AVS, which hinders result reproduction and comparison, we still employ AVSBench as our baseline for comparison.

~ \\ 
\noindent \textbf{Comparison on AVS:} From \cref{fig:case_appendix}, it is evident that our segmentation results show a significant improvement compared to AVSBench, demonstrating a substantial enhancement of audio-visual grounding.

~ \\ 
\noindent \textbf{Comparison on AVSS:} From \cref{fig:case_avss_appendix}, we can observe that our model achieves great performance gains on AVSS. Besides the noticeable improvements in semantic segmentation results, it is crucial to remind that our approach tries to mitigate the segmentation preference of audible objects. 

\section{Effectiveness of Audio Control}
\label{sec:audio_control_sup}

\begin{figure*}[tb] 
  \centering
   \includegraphics[width=0.8\linewidth]{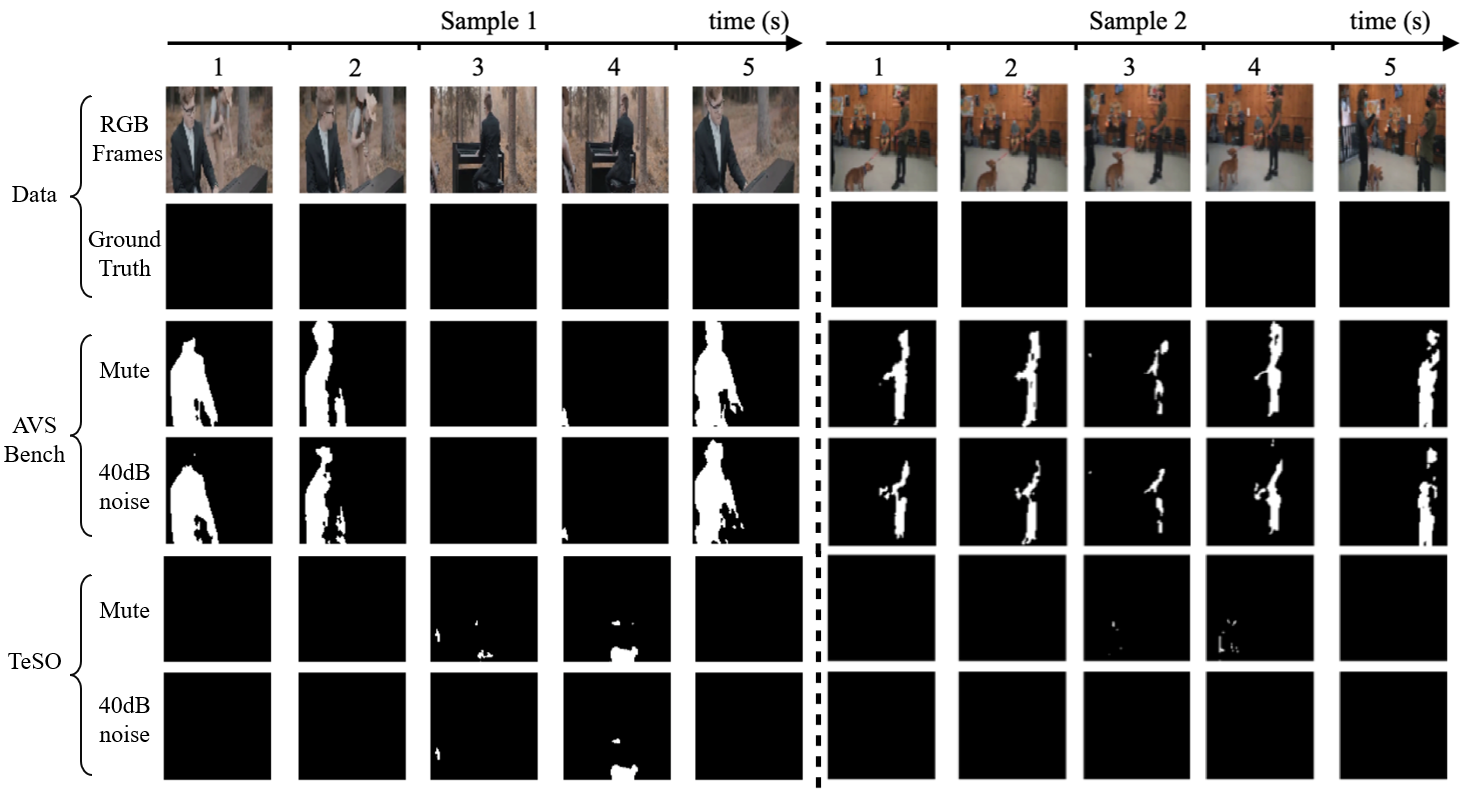}
   \caption{Comparison of audio control on AVS results with muted and 40dB pure White Gaussian Noise (WGN) audio input. The ground truth labels are all blank because the audio input is manually set wrong with muted or noise-only audio.}
   \label{fig:audio_control_appendix}
\end{figure*}

More examples of muted or noise-only audio are provided here to show the effectiveness of audio control. From \cref{fig:audio_control_appendix}, it is evident that our segmentation results achieved by TeSO exhibit almost entirely blank masks, indicating effective ``guidance'' by the muted or noise-only audio. As we mentioned in Sec. 4.4, our experiments reveal that the comparable results of some methods may to a certain extent rely on the segmentation preference of audible objects built during the training because they still segment audible objects (as shown in \cref{fig:audio_control_appendix}) and get comparable results even with pure Gaussian white noise.

\section{Effectiveness of Text Guidance}
\begin{figure*}[tb]
  \centering
   \includegraphics[width=0.8\linewidth]{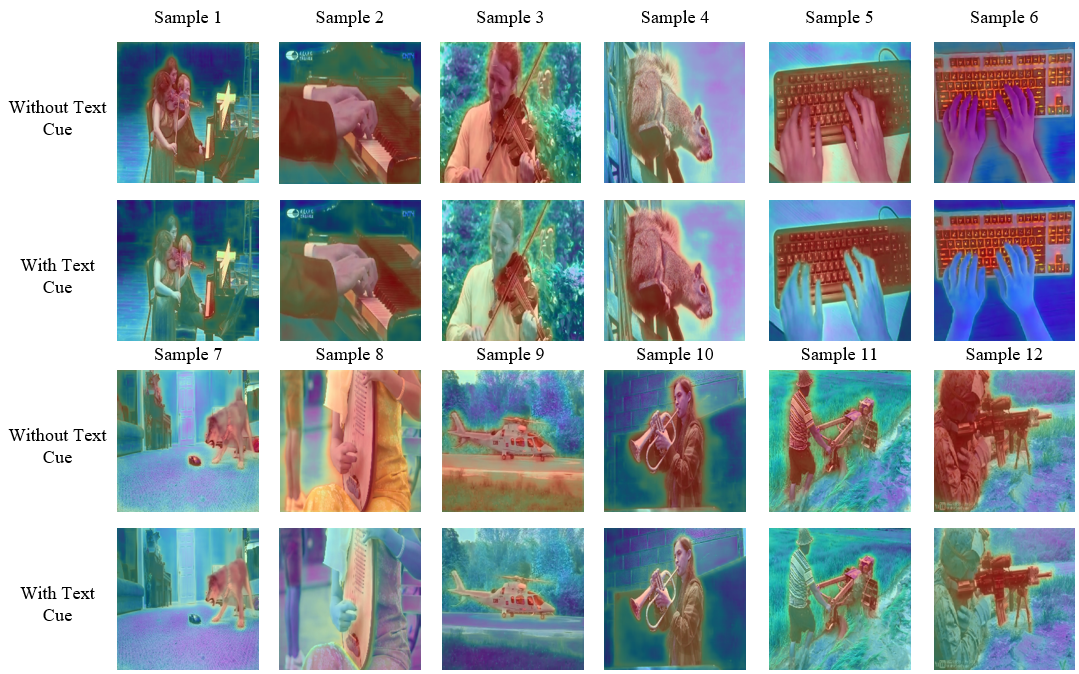}
   \caption{Comparison of whether text cues are used in TeSO. TeSO enhances audio-visual correlation by using text cues as a bridge between audio and visual modalities. Brighter (or redder) colors indicate the model's increased attention to these pixels.}
   \label{fig:textcue_appendix}
\end{figure*}

As shown in \cref{fig:textcue_appendix}, we present the qualitative results of introducing text cues to enhance the audio-visual correlation using a heatmap instead of the final binary mask, to show the degree of the model's attention on every pixel. By employing language as a bridge, we elevate audio-visual correlation by leveraging text cues as guidance, thereby enabling the model to better perceive the audio guidance.

\begin{table}[b]
\centering
\caption{Captioner and reasoner ablation on V1M-Test. The cooperation of LLaVA-1.5 and LLaMA-2 (the first column) works best.}
\label{tab:llm_ablation}
\resizebox{1\textwidth}{!}{%
\begin{tabular}{lccccc}
\hline
 & (a) & (b) & (c) & (d) & (e) \\ \hline
Captioner & LLaVA-1.5-7b & LLaVA-1.5-7b & VideoLLaMA-7b & LLaVA-1.5-7b & / \\
Reasoner & LLaMA-2-7b & LLaVA-1.5-7b & LLaMA-2-7b & NLTK & Manual video-level noisy label \\ \hline
mIoU & \textbf{66.02} & 65.41 & 64.27 & 63.87 & 65.53 \\
F-score & \textbf{0.801} & 0.790 & 0.788 & 0.783 & 0.792 \\ \hline
\end{tabular}%
}
\end{table}

\section{Ablation of Captioner and Reasoner}
As shown in \cref{tab:llm_ablation},
(a) and (b) indicate that using a text-only LLM LLaMA-2 as a reasoner outperforms a multimodal LLM LLaVA-1.5.
(a) and (c) demonstrate that LLaVA-1.5 is more effective as a frame-level dense captioner.
(a) and (d) show that benefits from the reasoning ability, LLM reasoner outperforms a naive NLTK noun parser.
(a) and (e) suggest that our frame-level reasoning approach is more accurate than using manual video-level noisy labels.

\section{Ablation of Audio Backbone.}
We use the BEATS$_{iter3}$ as an alternative to extract audio features. 
As shown in \cref{tab:beats}, we achieved a modest improvement of 0.24\%. However, we use VGGish for a fair comparison with most of the other methods.

\begin{table}[tb]

\centering
\caption{Ablation of audio backbone on V1M-Test. Model with BEATs audio feature gets better performance.}
\label{tab:beats}
\resizebox{0.45\textwidth}{!}{%

\begin{tabular}{cccc}

\hline
Method & Backbone & mIoU(\%)  & Fscore    \\ \hline
BAVS                    & BEATs                     & 59.63       & 0.659     \\
Ours                    & VGGish                    & 66.02       & 0.801     \\
Ours                    & BEATs                     & \textbf{66.26}       & \textbf{0.803}     \\ \hline
\end{tabular}%
}
\end{table}

\section{Ablation of Different Prompts}
We experienced with different prompt templates to explore the best prompting method. The alternative methods are divided into three categories based on instructive (No.1-6), misleading (No.7-9), and irrelevant (No.10) effects on segmentation, as shown in Tab. \ref{tab:prompt_template}.

\begin{table*}[tb]
\centering
\caption{
    Text assistance and prompt template. We implement diverse prompt templates on LLaMA-2-7b-Q4 to generate semantic text labels with varying levels of quality in V1M. Subsequently, we assess the text's capability to control the segmentation results when subjected to different quality conditions of semantic text labels.
} 
\scalebox{0.7}{
\begin{tabular}{llcllll}
\hline
No. & Category                     & CoT & Few shot  & Template & mIoU(\%) & F-score \\ \hline
1   & \multirow{6}{*}{instructive} & \ding{52} & few-shot  & $N \times$ templates shown in Fig. 3.        & \textbf{66.02} & \textbf{0.801}   \\
2   &                              & \ding{52} & one-shot  & $1 \times$ template shown in Fig. 3.        & 65.33  & 0.792  \\
3   &                              & \ding{52} & zero-shot & ``Let's think step by step to obtain sounding objects in the caption.''               & 64.76 & 0.787   \\
4   &                              & \ding{56}  & few-shot  & $N \times$ questions of sounding objects with direct answers.       & 64.80 & 0.785   \\
5   &                              & \ding{56}  & one-shot  & $1 \times$ question of sounding objects with direct answer.       & 64.08  & 0.773  \\
6   &                              & \ding{56}  & zero-shot & ``Please tell me the sounding objects in the caption.''       & 63.83  & 0.775  \\ \hline
7   & \multirow{3}{*}{misleading}  & \ding{56}  & zero-shot &  ``Please tell me the most possible sounding object in the caption.''      & 62.98  & 0.753  \\
8  &                              & \ding{56}  & zero-shot & ``Please tell me any objects in the caption.''       & 63.39  & 0.765 \\
9   &                              & \ding{56}  & zero-shot & ``Please tell me the background objects in the caption.''       & 61.52  & 0.740  \\ \hline
10   & \multirow{1}{*}{irrelevant}  & \ding{56}  & zero-shot & ``Tell me any random object.''       & 60.86  & 0.722 \\\hline
\end{tabular}
}
 
\label{tab:prompt_template}
\end{table*}

The inductive approach involves employing various prompt methods to acquire informative guidance. By examining the impact of this instructive information on segmentation, we can analyze the significance of incorporating CoT within the method. Utilizing a few-shot template with CoT can lead to a performance improvement of up to 1.22\% (No.1 vs. No.4). Furthermore, we devise an effective few-shot (No.1) template that yields a performance improvement of up to 1.25\% compared to the zero-shot (No.3) inference process. Finally, our method observes that the combination of few-shot templates and CoT leads to optimal performance.

In a similar context, we conduct experiments to investigate the impact of misleading text labels on the model. Specifically, we explore the effects of missing (No.7), ambiguous (No.8), and erroneous (No.9) text semantic information on the model's performance. The presence of such problematic semantic information results in a significant 4.50\% decrease in mIoU. Moreover, the introduction of irrelevant information (No.10) causes the largest drop in performance, amounting to 5.16\% performance.

\section{Multi-instance Scenario}
The multi-instance scenario (\eg, two men in the same frame) is a challenging problem in the field of AVS, especially since the existing mono-channel audio input cannot provide sufficient audio spatial information to the model. While our focus in this work is on the segmentation preference problem caused by weak audio guidance, we believe that our method has the potential to address such issues because visual scene descriptions can provide guidance that is difficult to obtain through audio features. For example, ``a man in red clothes on the right is singing a song while the man in black is smiling'' can help eliminate the ambiguity in multi-instance scenes. In future work, we hope to explore the multi-instance problem in AVS using our framework.

\begin{table*}[!t]
\centering
\caption{Performance on AVS-V3 for testing the generalization on unseen objects.
}
\scalebox{1}{
\begin{tabular}{lccccccccccc}
\hline
& \multicolumn{2}{c}{0-shot} & \multicolumn{2}{c}{1-shot} & \multicolumn{2}{c}{3-shot} & \multicolumn{2}{c}{5-shot} \\ 

Method & mIoU(\%)     & F-score  & mIoU(\%)     & F-score   & mIoU(\%)     & F-score     & mIoU(\%)     & F-score   \\ 

\hline
AVSBench  & 53.00 & 0.707 & 56.11 & 0.754 & 63.22 & 0.767 & 63.87 & 0.783 \\

AVSegFormer & 54.26 & 0.715 & 58.30 & 0.764 & 64.19 & 0.774 & 65.17 & 0.785 \\ 

GAVS & 54.71 & 0.722 & 62.89 & 0.768  & 66.28 & 0.774 & 67.75 & 0.795 \\ 

TeSO (ours) & \textbf{61.06} & \textbf{0.743} & \textbf{65.41} & \textbf{0.775}  & \textbf{69.29} & \textbf{0.791} & \textbf{72.15} & \textbf{0.810} \\

\hline
\end{tabular}
}

\label{tab:avs-v3}
\end{table*}

\section{Generalization}
Benefiting from the strong generalization of LLMs, it is possible to perform visual scene understanding on almost any scene and infer potential sounding objects as text cues. Therefore, we explore whether the model can exhibit better segmentation generalization performance for unseen object categories with the assistance of generalizable text cues from LLMs. As shown in \cref{tab:avs-v3}, our experiments on the AVS-V3 dataset demonstrate that our approach can significantly improve the generalization performance compared to other methods. This improvement may be attributed to the generalizable text cues reasoned by LLMs from scene descriptions, establishing a better audio-visual correlation.

\section{Versatility} 
\begin{table*}[!t]
    \centering
        
    \caption{Performance of TeSO on AVS-Benchmarks with different visual foundation models. TeSO-M2F uses Mask2Former as the visual foundation model while TeSO-SAM uses SAM as the visual foundation model.}
    \begin{threeparttable}
    \resizebox{1\textwidth}{!}{
    \begin{tabular}{lcccccccc}
        \hline
        \multirow{2}{*}{Method} & \multirow{2}{*}{Audio-backbone} & \multirow{2}{*}{Visual-backbone} & \multicolumn{2}{c}{V1S} & \multicolumn{2}{c}{V1M} & \multicolumn{2}{c}{AVSS-binary}  \\ 
         &  &  & mIoU(\%) & F-score & mIoU(\%) & F-score & mIoU(\%) & F-score \\ 
    
        \hline
        AVSBench & VGGish & PVT-v2 & 78.70 & 0.879 & 54.00 & 0.645 &  62.45 & 0.756 \\ 

        BAVS & Beats & Swin-Base & 82.68 & 0.898 & 59.63 & 0.659   & 55.45 & 0.640 \\


        TeSO-SAM (ours) & VGGish & ViT-Base & 82.68 & 0.903 & 64.97 & 0.794   & 68.16 & 0.808 \\ 

        TeSO-M2F (ours) & VGGish & Swin-Base & \textbf{82.84} & \textbf{0.917} & \textbf{66.02} & \textbf{0.801}   & \textbf{68.53} & \textbf{0.813} \\ 

        \hline
    \end{tabular}
    }
    \end{threeparttable}

    \label{tab:sam-avs}
\end{table*}
Our text-guided method is generally applicable to various visual foundation models with a transformer-based decoder. As depicted in \cref{tab:sam-avs}, we present the outcomes obtained by employing two different popular visual foundation models, namely SAM (Segment Anything Model) and Mask2Former, in their base versions. Even when using SAM as the foundation model, our method remains effective and produces comparable results.
 
%
\end{document}